\documentclass[10pt,twocolumn,letterpaper]{article}

\usepackage{forloop}
\usepackage{etoolbox}

\usepackage[table]{xcolor}
\usepackage{titlesec}

\usepackage{subcaption}
\usepackage{forloop}
\usepackage{etoolbox}

\usepackage{caption}
\newcommand{\cN}{\mathcal{N}}

\newcommand{\cS}{\mathcal{S}}

\newcommand{\cX}{\mathcal{X}}

\newcommand{\cY}{\mathcal{Y}}

\newcommand{\cV}{\mathcal{V}}
\newcommand{\cU}{\mathcal{U}}

\def\exp{\operatorname*{exp\,}}
\def\log{\operatorname*{log\,}}

\def\btheta{{\boldsymbol \theta}}

\def\x{{\boldsymbol x}}

\def\y{{\boldsymbol y}}

\def\gradient{{\nabla}}
\def\expect{{\mathbb E}}

\def\Prr{{\displaystyle P}}

\newcommand{\comment}[1]{}

\newcommand{\fnorm}[2][2]{\ensuremath{ \left\| #2 \right\|^2_{ \mathrm{#1} } } }

\def\x{{\boldsymbol x}}

\newcommand*\rot{\rotatebox{90}}

\def\best{\bf \cellcolor[gray]{0.85}}

\titlespacing\paragraph{0pt}{5pt}{5pt}

\expandafter\def\expandafter\normalsize\expandafter{%
\normalsize\setlength\abovedisplayskip{3pt}}

\expandafter\def\expandafter\normalsize\expandafter{%
\normalsize\setlength\belowdisplayskip{3pt}}

\usepackage{cvpr}
\usepackage{times}
\usepackage{epsfig}
\usepackage{graphicx}
\usepackage{amsmath}

\usepackage{amssymb, eucal}

\usepackage{fancyhdr}
\usepackage[pagebackref=true,breaklinks=true,letterpaper=true,colorlinks,bookmarks=false]{hyperref}

\cvprfinalcopy %

\def\cvprPaperID{1140} %

\begin{document}

\title{Efficient Piecewise Training of Deep Structured Models  for Semantic Segmentation}

\author{Guosheng Lin, \;  Chunhua Shen, \;
Anton van den Hengel, \; Ian Reid\\
The University of Adelaide; and Australian Centre for Robotic Vision\\
}

\maketitle
\begin{abstract}
   Recent advances in  semantic image segmentation have mostly been achieved by
  training deep convolutional neural networks (CNNs).
  We show how to improve semantic segmentation through the use of contextual information;
	specifically, we explore `patch-patch' context between image regions, and `patch-background' context.
  For learning from the patch-patch context,
 we formulate Conditional Random Fields (CRFs) with CNN-based pairwise potential functions to capture semantic correlations between neighboring patches.
  Efficient piecewise training of the proposed deep structured model is then applied to avoid repeated expensive CRF inference
  for back propagation.
   For capturing the patch-background context, we show that a network design
   with traditional multi-scale image input and sliding pyramid pooling is effective for improving performance.
	Our experimental results set new state-of-the-art performance on a number of
    popular semantic segmentation datasets, including NYUDv2, PASCAL VOC 2012, PASCAL-Context, and SIFT-flow.
	In particular, we achieve an intersection-over-union score of $78.0$ on the challenging PASCAL VOC 2012 dataset.
\end{abstract}

\section{Introduction}

Semantic image segmentation aims to predict a category label for every image pixel,
which is an important yet challenging task for image understanding.
Recent approaches have applied convolutional neural network (CNNs) \cite{farabet2013learning,LongSD14,ChenPKMY14}
to this pixel-level labeling task and achieved remarkable success.
Among these CNN-based methods, fully convolutional neural networks (FCNNs)~\cite{LongSD14,ChenPKMY14}
have become a popular choice, because of their computational efficiency for dense prediction and end-to-end style learning.

Contextual relationships are ubiquitous and provide important cues for scene understanding tasks.
Spatial context can be formulated in terms of  semantic compatibility relations between one object and its neighboring objects or image patches (stuff), in which a compatibility relation is an indication of the co-occurrence of visual patterns.
For example, a car is likely to appear over a road, and a glass is likely to appear over a table.
Context can also encode incompatibility relations.
For example, a car is not likely to be surrounded by sky.  These relations also exist
at finer scales, for example, in
object part-to-part relations, and part-to-object relations.
In some cases, contextual information is the most important cue, particularly when a single object shows significant visual ambiguities.
A more detailed discussion of the value of spatial context can be found in \cite{heitz2008learning}.

\begin{figure}[t]
	\centering
	\includegraphics[width=.95\linewidth]{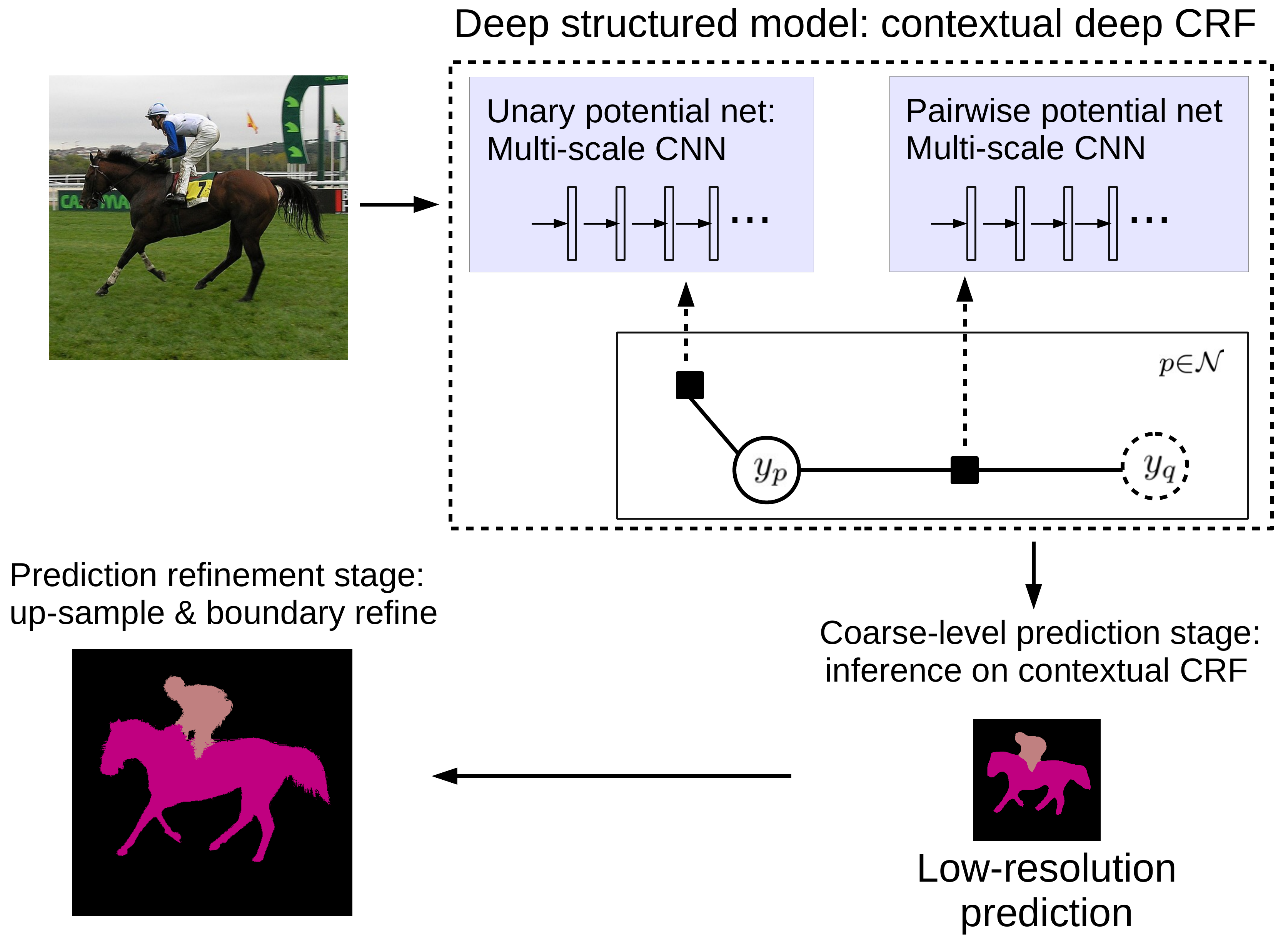}
\caption{An illustration of the prediction process of our method.
Both our unary and pairwise potentials are formulated as multi-scale CNNs for capturing semantic relations between image regions.
Our method outputs low-resolution prediction after CRF inference, then the prediction is up-sampled and refined in a standard post-processing stage to output the final prediction.}
\label{fig:general_graph}
\end{figure}

We explore two types of spatial context to improve the segmentation performance: patch-patch context and patch-background context.
The patch-patch context is the semantic relation between the visual patterns of two image patches. Likewise,
patch-background context is the semantic relation between a patch and a large background region.

Explicitly modeling the patch-patch contextual relations has not been well studied in recent CNN-based segmentation methods.
In this work, we propose to explicitly model the contextual relations using conditional random fields (CRFs).
We formulate CNN-based pairwise potential functions to capture semantic correlations between neighboring patches.
Some recent methods  combine CNNs and CRFs for semantic segmentation,
e.g., the dense CRFs applied in \cite{ChenPKMY14,schwing2015fully,zheng2015conditional,Dai2015arXiv}.
The purpose of applying the dense CRFs in these methods is to refine the upsampled low-resolution prediction to sharpen object/region boundaries.
These methods consider Potts-model-based pairwise potentials for enforcing local smoothness.
There the pairwise potentials are conventional log-linear functions.
In contrast, we
learn more general pairwise potentials using CNNs to model the semantic compatibility between image regions.
Our CNN pairwise potentials aim to improve the coarse-level prediction rather than doing local smoothness, 
and thus have a different purpose compared to Potts-model-based pairwise potentials.
Since these two types of potentials have different effects, they can be combined to improve the segmentation system.
Fig.~\ref{fig:general_graph} illustrates our prediction process.

In contrast to patch-patch context,
patch-background context is widely explored in the literature.
For CNN-based methods,
background information can be effectively captured
by combining features from a multi-scale image network input, and has shown good performance
in some recent segmentation methods \cite{farabet2013learning,MostajabiYS14}.
A special case of capturing patch-background context is considering the whole image as the background region and incorporating the image-level label information into learning.
In our approach, to encode rich background information, we construct multi-scale networks and apply sliding pyramid pooling on feature maps.
The traditional pyramid pooling (in a sliding manner) on the feature map is able to capture information from background regions of different sizes.

Incorporating general pairwise (or high-order) potentials usually involves expensive inference, which brings challenges for CRF learning.
To facilitate efficient learning we apply piecewise training of the CRF \cite{SuttonM05} to avoid repeated inference
 during back propagation training.

Thus our main \emph{contributions} are as follows.

  { 1.}~We formulate CNN-based general pairwise potential functions in CRFs to explicitly model patch-patch semantic relations.

  { 2.}~Deep CNN-based general pairwise potentials are challenging for efficient CNN-CRF joint learning.
  We perform approximate training, using piecewise training of CRFs \cite{SuttonM05}, 
  to avoid the repeated inference at every stochastic gradient descent iteration and thus achieve efficient learning.

  {3.}~We explore background context by applying a network architecture with traditional multi-scale image input \cite{farabet2013learning} 
  and sliding pyramid pooling \cite{lazebnik2006beyond}.
   We empirically demonstrate the effectiveness of this network architecture for semantic segmentation.

  {4.}~We set new state-of-the-art performance on a number of popular semantic segmentation datasets, including NYUDv2, PASCAL VOC 2012, PASCAL-Context, and SIFT-flow.
	In particular, we achieve an intersection-over-union score of $78.0$ on the PASCAL VOC 2012 dataset, which is {\em  the  best reported result} to date.

\subsection{Related work}

Exploiting contextual information
has been widely studied in the literature (e.g., \cite{rabinovich2007objects,heitz2008learning,doersch2014context}).  
For example, the early work ``TAS'' \cite{heitz2008learning} models different types of spatial context between {\em Things} and {\em Stuff}
using a generative probabilistic graphical model.

The most successful recent methods for semantic image segmentation are based on CNNs.
A number of these CNN-based methods for segmentation are region-proposal-based methods \cite{GirshickDDM13,BharathECCV2014}, which first generate region proposals and then assign category labels to each.  Very recently,  FCNNs
 \cite{LongSD14,ChenPKMY14,Dai2015arXiv} have become a popular choice for semantic segmentation, because of their effective feature generation and end-to-end training.
FCNNs
have also been applied to a range of other dense-prediction tasks recently,
 such as image restoration \cite{Eigen_iccv13}, image super-resolution \cite{Dong_eccv14}
 and depth estimation \cite{dcnn_nips14,liu2014deep}.
The method we propose here is similarly
built upon fully convolution-style networks.

The direct prediction of FCNN based methods usually are in low-resolution.
To obtain high-resolution predictions, a number of recent methods focus on refining the low-resolution prediction to obtain high resolution prediction.
DeepLab-CRF \cite{ChenPKMY14} performs bilinear upsampling of the prediction score map to the input image size and apply the dense CRF method \cite{krahenbuhl2012efficient} to refine the object boundary by leveraging the color contrast information.
CRF-RNN \cite{zheng2015conditional} extends this approach by implementing recurrent layers for end-to-end learning of the dense CRF and the FCNN network.
The work in \cite{noh2015learning} learns deconvolution layers to upsample the low-resolution predictions.
The depth estimation method \cite{liu2015learning} explores super-pixel pooling 
for building the gap between low-resolution feature map and high-resolution final prediction.
Eigen \textit{et~al.} \cite{eigen2015predicting} perform coarse-to-fine learning of multiple networks with different resolution outputs for refining the coarse prediction.
The methods in \cite{hariharan2014hypercolumns,LongSD14} explore middle layer features (skip connections) for high-resolution prediction.
Unlike these methods, our method focuses on improving the coarse (low-resolution) prediction 
by learning general CNN pairwise potentials to capture semantic relations between patches.
These refinement methods are complementary to our method.

Combining the strengths of CNNs and CRFs for segmentation
has been the focus of several recently developed approaches.
DeepLab-CRF in \cite{ChenPKMY14} trains FCNNs and applies a dense CRF~\cite{krahenbuhl2012efficient} method
as a post-processing step.
CRF-RNN \cite{zheng2015conditional} and the method in \cite{schwing2015fully} extend DeepLab and \cite{krahenbuhl2013parameter} by jointly learning the dense CRFs and CNNs.  They consider Potts-model based  pairwise potential functions which  enforce smoothness only.
The CRF model in these methods is for refining the up-sampled prediction.
Unlike  these methods, our approach learns CNN-based pairwise potential functions for modeling semantic relations between patches.

Jointly learning CNNs and CRFs has also been explored in other applications apart from segmentation.
The recent work in \cite{liu2014deep,liu2015learning} proposes to jointly learn {\em continuous} CRFs and CNNs for depth estimation
from  single monocular images. 
The work in \cite{Lecun_nips14} combines CRFs and CNNs for human pose estimation.
The authors of \cite{chen2014learning} explore joint training of Markov random fields and deep neural networks for predicting words from noisy images and image s classification. Different from these methods, we explore efficient piecewise training of CRFs with CNN pairwise potentials.

\begin{figure}[t]
	\center
	\includegraphics[width=.8\linewidth]{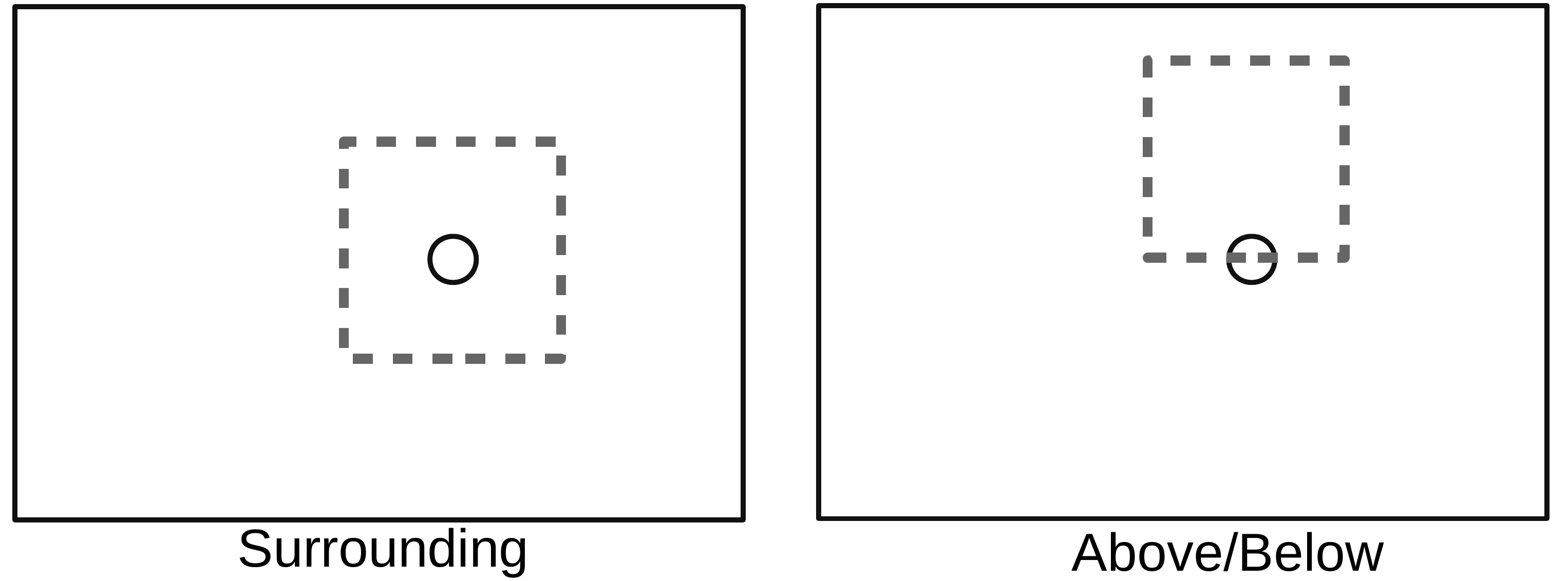}
\caption{An illustration of constructing pairwise connections in a CRF graph.
A node is connected to all other nodes which lie within the range box (dashed box in the figure).
Two types of spatial relations are described in the figure, which
correspond to two types of pairwise potential functions.
}
\label{fig:pairwise}
\end{figure}

\section{Modeling semantic pairwise relations}

\begin{figure}[t]
	\centering
	\includegraphics[width=1\linewidth]{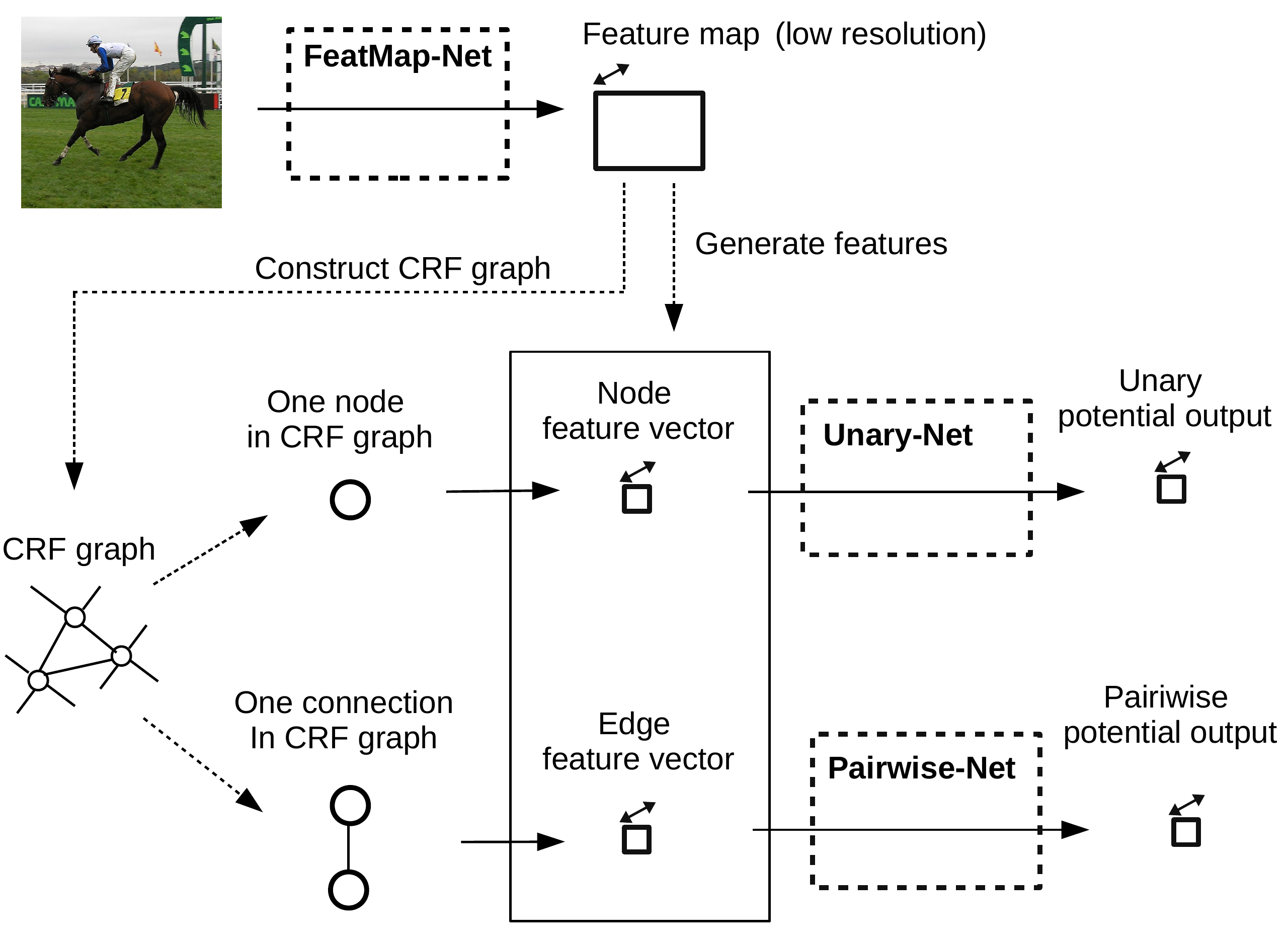}
\caption{An illustration of generating unary or pairwise potential function outputs.
First a feature map is generated by a FeatMap-Net, and a CRF graph is constructed based on the spatial resolution of the feature map.
Finally the Unary-Net (or Pairwise-Net) produces potential function outputs.}
\label{fig:potential_net}
\end{figure}

Fig.~\ref{fig:potential_net} conceptualizes our architecture at a high level.
Given an image, we first apply a convolutional network to generate a feature map.
We refer to this network as `FeatMap-Net'.
The resulting feature map is at a lower resolution than the original image because of the down-sampling operations in the pooling layers.

We then create the CRF graph as follows:
for each location in the feature map (which corresponds to a rectangular region in the input image) we create one node in the CRF graph.
Pairwise connections in the CRF graph are constructed by connecting one node to all other nodes 
which lie within a spatial range box (the dashed box in Fig.~\ref{fig:pairwise}).
We consider different spatial relations by defining different types of range box, and each type of spatial relation is modeled by a specific pairwise potential function.
As shown in Fig.~\ref{fig:pairwise},  our method models the ``surrounding" and ``above/below" spatial relations.
In our experiments, the size of the range box (dash box in the figure) size is $0.4a \times 0.4a$.
Here we denote by $a$ the length of the short edge of the feature map.

Note that although `FeatMap-Net' defines a common architecture, in fact we train three such networks: one for the unary potential and one each for the two types of pairwise potential.

\section{Contextual Deep CRFs}
\label{sec:crf_details}

Here we describe the details of our deep CRF model.
We denote by $\x \in \cX$ one input image and $\y \in \cY$ the labeling mask which describes the label configuration of each node in the CRF graph.
The energy function is denoted by $E(\y, \x; \btheta)$ which models the compatibility of the input-output pair, with a small output value indicating high confidence in the prediction $\y$.
All network parameters are denoted by $\btheta$ which we need to learn.
The conditional likelihood for one image is formulated as follows:
\begin{equation}\label{eq:prob}
\begin{aligned}
\small
\Prr(\y|\x) = \frac{1}{Z(\x)} \exp [- E(\y, \x)].
\end{aligned}
\end{equation}
Here $Z(\x) = \sum_{\y} \exp [ -E(\y, \x) ]$ is the partition function.
The energy function is typically formulated by a set of unary and pairwise potentials:
\begin{align}
\label{eq:energy}
\small
	E(\y, \x) = &  \sum_{U \in \cU} \sum_{p \in \cN_U } U(y_{p}, \x_p) %
  + \sum_{V \in \cV} \sum_{(p,q) \in \cS_V} V(y_{p}, y_{q}, \x_{pq}). \notag
\end{align}
Here $U$ is a unary potential function,
and to make the exposition more general, we consider multiple types of unary potentials with
$\cU$ the set of all such unary potentials.
$\cN_U$ is a set of nodes for the potential $U$.
Likewise, $V$ is a pairwise potential function
 with $\cV$ the set of all types of pairwise potential. $\cS_V$ is the set of edges for the potential $V$.
$\x_{p}$ and $\x_{pq}$  indicates the corresponding image regions which associate to the specified node and edge.

\begin{figure*}[t]
	\center
	\includegraphics[width=.9\linewidth]{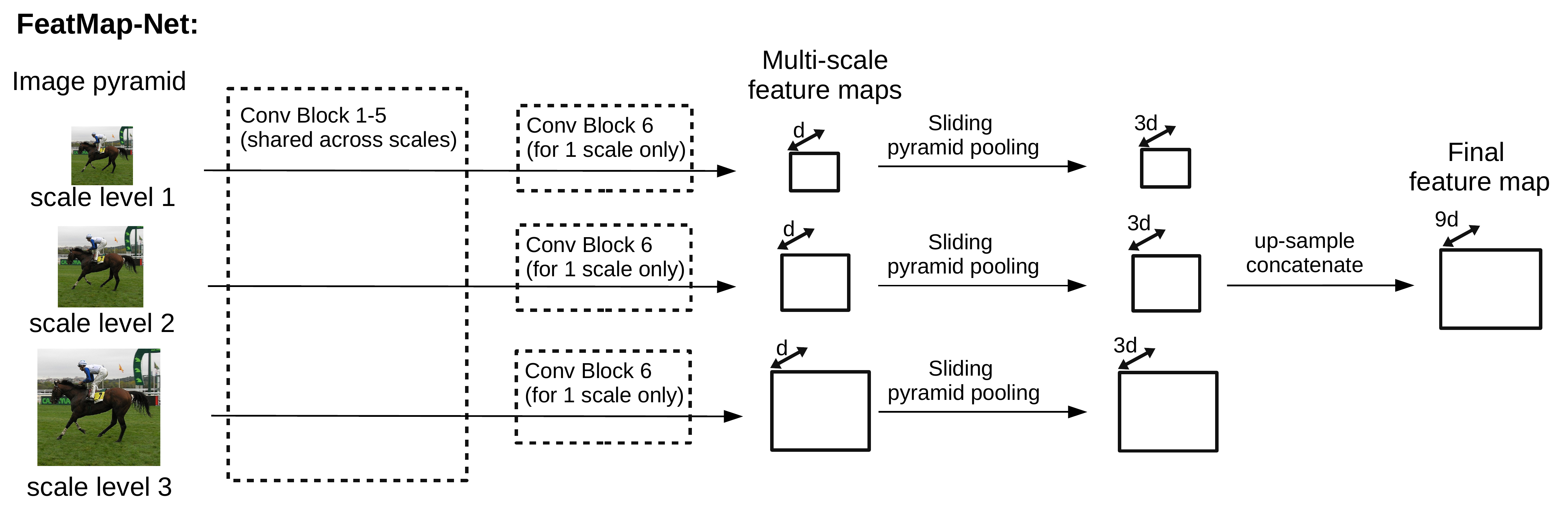}
\caption{The details of our FeatMap-Net.
An input image is first resized into $3$ scales,
then each resized image goes through 6 convolution blocks to output one feature map.
Top $5$ convolution blocks are shared for all scales. Every scale has a specific convolution block (Conv Block 6).
We perform $2$-level sliding pyramid pooling (see Fig.~\ref{fig:pooling} for details).
$d$ indicates the feature dimension.
}
\label{fig:featmapnet}
\end{figure*}

\subsection{Unary potential functions}
We formulate the unary potential function by stacking the FeatMap-Net for generating feature maps and 
a shallow fully connected network (referred to as Unary-Net) to generate the final output of the unary potential function.
The unary potential function is written as follows:
\begin{equation}
	U(y_{p}, \x_p; \btheta_U) = - z_{p,y_p} (\x; \btheta_U).
\end{equation}
Here $z_{p, y_p }$ is the output value of Unary-Net, which corresponds to the $p$-th node and the $y_p$-th class.

Fig.~\ref{fig:potential_net} includes an illustration of the Unary-Net and how it integrates with FeatMap-Net.
The unary potential at each CRF node is simply the $K$-dimensional output (where $K$ is the number of classes) of Unary-Net applied to the node feature vector from the correpsonding location in the feature map (i.e. the output of FeatMap-Net).

\subsection{Pairwise potential functions}
Fig. \ref{fig:potential_net} likewise illustrates how the pairwise potentials are generated.  The edge features are formed by concatenating 
the corresponding feature vectors of two connected nodes (similar to~\cite{kolesnikov2014closed}).  The feature vector for each node in the pair is from the feature map output by FeatMap-Net. The edge features of one pair  are then fed to a shallow fully connected network (referred to as Pairwise-Net) to generate the final output that is the pairwise potential.  The size of this is $K \times K$ to match the number of possible label combinations for a pair of nodes. 
The pairwise potential function is written as follows:
\begin{align}
	V & (y_{p}, y_{q},  \x_{pq}; \btheta_V) = - z_{p, q, y_p, y_q} (\x; \btheta_V).
\end{align}
Here $z_{p, q,  y_p,  y_q }$ is the output value of Pairwise-Net.
It is the confidence value for the node pair $(p, q)$ when they are labeled with the class value $( y_p,  y_q )$, which measures the compatibility of the label pair $(y_{p}, y_{q}$) given the input image $\x$.
$\btheta_V$ is the corresponding set of CNN parameters for the potential $V$, which we need to learn.

Our formulation of pairwise potentials is different from the Potts-model-based formulation in the existing methods of \cite{ChenPKMY14,zheng2015conditional}.
The Potts-model-based pairwise potentials are a log-linear functions and
employ a special formulation for enforcing neighborhood smoothness.
In contrast, our pairwise potentials model the semantic compatibility between two nodes
 with the output for every possible value of the label pair $(y_{p}, y_{q}$) individually parameterized by CNNs.

In our system, after obtaining the coarse level prediction, we still need to perform a refinement step to obtain the final high-resolution prediction 
(as shown in Fig.~\ref{fig:general_graph}).
Hence we also apply the dense CRF method \cite{krahenbuhl2012efficient}, as in many other recent methods, in the prediction refinement step.
Therefore, our system takes advantage of both contextual CNN potentials and the traditional smoothness potentials to improve the final system.
More details are described in Sec.~\ref{sec:prediction}.

As in \cite{winn2006layout,heesch2010markov},
modeling asymmetric relations requires the potential function is capable of modeling input orders, 
since we have: $ V  (y_{p}, y_{q},  \x_{pq}) \neq V (y_{q}, y_{p},  \x_{qp})$.
Take the asymmetric relation ``above/below" as an example;
we take advantage of the input pair order to indicate the spatial configuration of two nodes,
thus the input $(y_p, y_q, \x_{pq})$
indicates the configuration that the node $p$ is spatially lies above the node $q$.

The asymmetric property is readily achieved with our general formulation of pairwise potentials.  
The potential output for every possible pairwise label combination for $(p,q)$ is individually parameterized by the pairwise CNNs.

\section{Exploiting background context}
\label{sec:network_details}

To encode rich background information,
we use multi-scale CNNs and  sliding pyramid pooling \cite{lazebnik2006beyond} for our FeatMap-Net.
Fig.~\ref{fig:featmapnet} shows the details of the FeatMap-Net.

CNNs with multi-scale image network inputs have shown good performance in some recent segmentation methods \cite{farabet2013learning,MostajabiYS14}.
The traditional pyramid pooling (in a sliding manner) on the feature map is able to capture information from background regions of different sizes.
We observe that these two techniques (multi-scale network design and pyramid pooling) 
for encoding background information are very effective for improving performance.

Applying CNNs on multi-scale images has shown good performance in some recent segmentation methods \cite{farabet2013learning,MostajabiYS14}.
In our multi-scale network, an input image is first resized into $3$ scales,
then each resized image goes through 6 convolution blocks to output one feature map.
In our experiment, the $3$ scales for the input image are set to $1.2$, $0.8$ and $0.4$.
All scales share the same top $5$ convolution blocks.
In addition, each scale has an exclusive convolution block (``Conv Block 6" in the figure) 
which captures scale-dependent information.
The resulting $3$ feature maps (corresponding to $3$ scales) are of different resolutions, therefore we  upscale the two smaller ones to the size of the largest feature map using bilinear interpolation. These feature maps are then concatenated to form one feature map.

We perform spatial pyramid pooling \cite{lazebnik2006beyond} (a modified version using sliding windows) 
on the feature map to capture information from background regions in multiple sizes.
This increases the field-of-view for the feature map and thus it is able to capture the information from a large image region.
Increasing the field-of-view generally helps to improve performance \cite{ChenPKMY14}.

\begin{figure}[t]
	\center
	\includegraphics[width=1\linewidth]{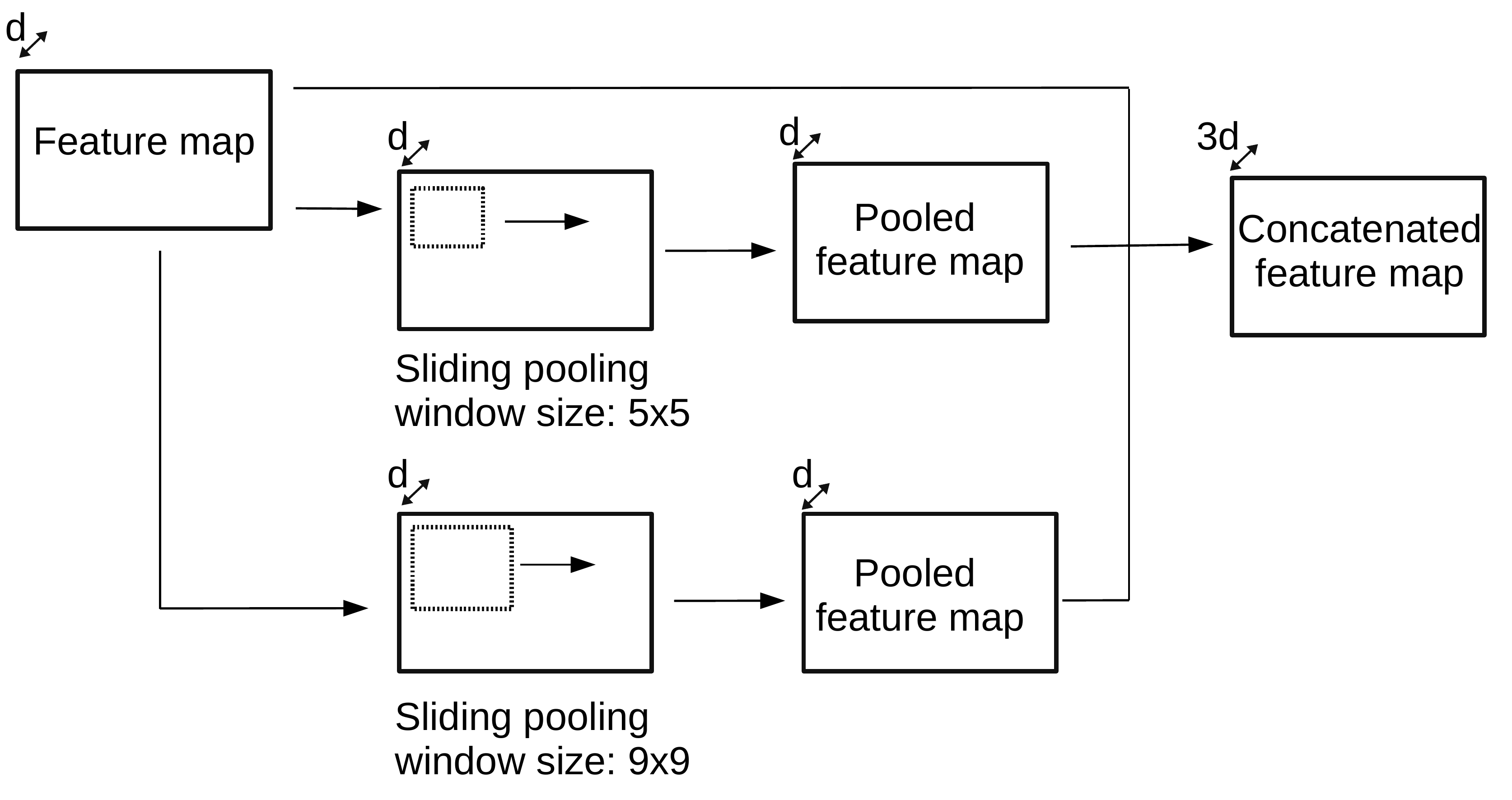}
\caption{
Details for sliding pyramid pooling.
We perform $2$-level sliding pyramid pooling on the feature map for capturing patch-background context, 
which encode rich background information and increase the field-of-view for the feature map.
}
\label{fig:pooling}
\end{figure}

The details of spatial pyramid pooling are illustrated in Fig. \ref{fig:pooling}.
In our experiment, we perform $2$-level pooling for each image scale.
We define $5 \times 5$ and $9 \times 9$ sliding pooling windows (max-pooling) to generate $2$ sets of pooled feature maps,
which are then concatenated to the original feature map to construct the final feature map.

\begin{figure}[t]
	\center
	\includegraphics[width=.75\linewidth]{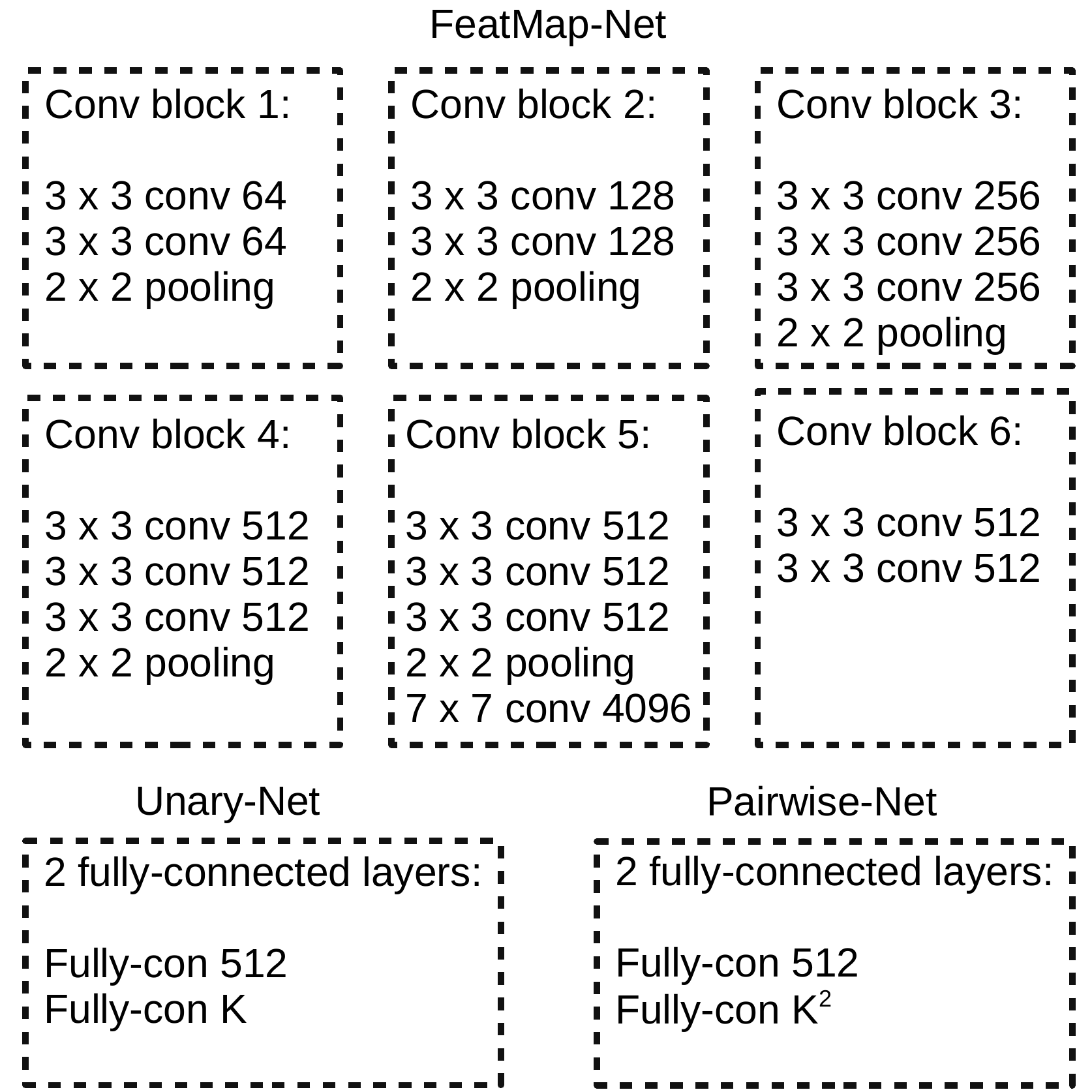}
\caption{The detailed configuration of the networks: FeatMap-Net, Unary-Net and Pairwise-Net.
  $K$ is the number of classes.
  For FeatMap-Net, the top 5 convolution blocks share the same configuration
as the convolution blocks in the VGG-16 network. The stride of the last max pooling layer is 1, and
for the other max pooling layers we use the same stride setting as VGG-16.
}
\label{fig:network_conf}
\end{figure}

The detailed network layer configuration for all networks are described in Fig.~\ref{fig:network_conf}.

\section{Prediction}
\label{sec:prediction}

In the prediction stage,
our deep structured model will generate low-resolution prediction (as shown in Fig.~\ref{fig:general_graph}), 
which is $1/16$ of the input image size.
This is due to the stride setting of pooling or convolution layers for sub-sampling.
Therefore, we apply two prediction stages for obtaining the final high-resolution prediction: 
the coarse-level prediction stage and the prediction refinement stage.

\subsection{Coarse-level prediction stage}
We perform CRF inference on our contextual structured model to obtain the coarse prediction of a test image.
We consider the marginal inference over nodes for prediction:
\begin{align} \label{eq:inference}
\forall p \in \cN: \;\; \Prr(y_p|\x)&= {\textstyle \sum_{\y \backslash y_p } \Prr(\y|\x)}.
\end{align}
The obtained marginal distribution can be further applied in the next prediction stage for boundary refinement.

Our CRF graph does not form a tree structure,
nor are the potentials submodular,
hence we need to an apply approximate inference.
To address this we
apply an efficient  message passing algorithm which is based on the mean field approximation \cite{Nowozinstruct}.
The mean field algorithm constructs a simpler distribution $Q(\y)$, e.g.,
a product of independent marginals: 
$Q(\y)=\prod_{p \in \cN} Q_p (y_p)$, 
which minimizes the KL-divergence between the distribution $Q(\y)$ and $\Prr(\y)$.
In our experiments, we perform $3$ mean field iterations.

\subsection{Prediction refinement stage}
We generate the score map for the coarse prediction from the marginal distribution which we obtain from the mean-field inference.
We first bilinearly up-sample the score map of the coarse prediction to the size of the input image.
Then we apply a common post-processing method \cite{krahenbuhl2012efficient} (dense CRF)
to sharpen the object boundary for generating the final high-resolution prediction.
This post-processing method leverages low-level pixel intensity information (color contrast) for boundary refinement. 
Note that most recent work on image segmentation similarly produces low-resolution prediction
 and have a upsampling and refinement process/model for the final prediction, e.g., 
\cite{ChenPKMY14,zheng2015conditional,Dai2015arXiv}.

In summary, we simply perform bilinear upsampling of the coarse score map and apply the boundary refinement post-processing.
We argue that this stage can be further improved by applying more sophisticated refinement methods, e.g., training deconvolution networks \cite{noh2015learning}, 
training multiple coarse to fine learning networks \cite{eigen2015predicting}, 
and exploring middle layer features for high-resolution prediction \cite{hariharan2014hypercolumns,LongSD14}.
It is expected that applying better refinement approaches will gain further performance improvement.

\section{CRF training}

A common approach for CRF learning is to maximize the likelihood,
or equivalently minimize the negative log-likelihood, which can be written
for one image as:
\begin{align}
 - \log \Prr(\y | \x; \btheta) = E(\y, \x; \btheta) + \log Z(\x; \btheta).
\label{eq:Prr}
\end{align}
Adding regularization to the CNN parameter $\btheta$, the optimization problem for CRF learning is: 
{\small
\begin{align}
\label{eq:crf_learning_org}
\min_{\btheta} \frac{\lambda}{2} \fnorm \btheta + \sum_{i=1}^N \biggr[ E(\y^{(i)}, \x^{(i)}; \btheta) + \log Z(\x^{(i)}; \btheta) \biggr].
\end{align}
}
Here $\x^{(i)}$, $\y^{(i)}$ denote the $i$-th training image and its segmentation mask; $N$ is the number of training images; $\lambda$ is the weight decay parameter.
We can apply stochastic gradient (SGD) based methods to optimize the above problem for learning $\btheta$.
The energy function $E (\y, \x; \btheta)$ is constructed from CNNs, and its
gradient $\gradient_\btheta E (\y, \x; \btheta)$ easily computed
by applying the chain rule as in conventional CNNs.
However, the partition function $Z$ brings difficulties for optimization.  Its gradient is:
\begin{align}
\gradient_\btheta  \log &  Z(\x ; \btheta)  \notag  \\
	 = &  \sum_\y \frac{\exp[-E(\y, \x; \btheta)]} {\sum_{\y'} \exp[-E(\y', \x; \btheta)]} \gradient_\btheta [ - E (\y, \x; \btheta)] \notag \\
	 = &  - \expect_{\y \sim \Prr(\y | \x ; \btheta)} \gradient_\btheta E (\y, \x; \btheta)
\end{align}
Generally the size of the output space $\cY$
is exponential in
the number of nodes, which prohibits the direct calculation of $Z$ and its gradient.
The CRF graph we considered for segmentation here is a loopy graph (not tree-structured),
for which the inference is generally computationally expensive.
More importantly, usually a large number of SGD iterations (tens or hundreds of thousands) are required for
training CNNs. Thus performing inference at each SGD iteration is very computationally expensive.

\subsection{Piecewise training of CRFs}

Instead of directly solving the optimization in \eqref{eq:crf_learning_org},
we propose to apply an approximate CRF learning method.
In the literature,
there are two popular types of learning methods which approximate the CRF objective
: pseudo-likelihood learning \cite{besag1977efficiency} and piecewise learning \cite{SuttonM05}.
The main advantage of these methods  in term of training deep CRF is that they  do not involve marginal inference for gradient calculation,
which significantly improves the efficiency of training.
Decision tree fields \cite{NowozinRBSYK11} and regression tree fields \cite{jancsary2012regression}
are based on pseudo-likelihood learning, while
piecewise learning has been applied in the work \cite{SuttonM05,kolesnikov2014closed}.

Here we develop this idea for the case of training the CRF with the CNN potentials.
In piecewise training, 
the conditional likelihood 
is formulated as a number of independent likelihoods defined on potentials, written as:
\begin{align}
\Prr (\y | \x) = \prod_{U \in \cU} \prod_{ p \in \cN_{U}} \Prr_{U} ( y_p | \x) \prod_{V \in \cV}
\prod_{ (p,q) \in \cS_{V}} \Prr_{V} ( y_p, y_q | \x). \notag
\end{align}
The likelihood $\Prr_{U} ( y_p | \x)$ is constructed from the unary potential $U$.
Likewise, $\Prr_{V} ( y_p, y_q | \x)$ is constructed from the pairwise potential $V$.
$\Prr_{U}$ and $\Prr_{V}$ are written as:
\begin{align}
\Prr_{U}(y_p|\x) &  = \frac{\exp [- U(y_p, \x_p)]}{\sum_{y'_p} \exp [ -U(y'_p, \x_p) ]}, \label{eq:pu} \\
\Prr_{V}(y_p, y_q|\x) &  = \frac{\exp [- V(y_p, y_q, \x_{pq})]}{\sum_{y'_p, y'_q} \exp [ -V(y'_p, y'_q, \x_{pq}) ]}
\label{eq:pv}.
\end{align}
 Thus the optimization for piecewise training is to minimize
 the negative log likelihood with regularization:
\begin{align}
	\min_{\btheta} \frac{\lambda}{2} & \fnorm \btheta -
	 \sum_{i=1}^N \biggr[ \sum_{U \in \cU}
		\sum_{ p \in \cN_{U}^{(i)}} \log \Prr_{U} ( y_p | \x^{(i)}; \btheta_U) \notag \\
	& + \sum_{V \in \cV} \sum_{ (p,q) \in \cS_{V}^{(i)}} \log \Prr_{V} ( y_p, y_q | \x^{(i)}; \btheta_V) \biggr].
	\label{eq:partitioned}
\end{align}
Compared to the objective in \eqref{eq:crf_learning_org} for direct maximum likelihood learning,
the above objective does not involve the global partition function $Z(\x ; \btheta)$.
To calculate the gradient of the above objective,
we only need to calculate the gradient
$\gradient_{\btheta_U} \log \Prr_{U}$
and
$\gradient_{\btheta_V} \log \Prr_{V}$.
With the definition in \eqref{eq:pu}, $\Prr_{U}$ is a conventional Softmax normalization function over only $K$ (the number of classes) elements. 
Similar analysis can also be applied to $\Prr_{V}$.
Hence, we can easily calculate the gradient without involving expensive inference.
Moreover, we are able to perform parallel training of potential functions, since the above objective is formulated as a summation of independent log-likelihoods.

 As previously discussed, CNN training usually involves a large number of gradient update iterations.  However this means that expensive inference during every gradient iteration becomes impractical.  Our piecewise approach here provides a practical solution for learning CRFs with CNN potentials on large-scale data.

\section{Experiments}

We evaluate our method on $4$ popular semantic segmentation datasets: PASCAL VOC 2012, NYUDv2, PASCAL-Context and SIFT-flow.
The segmentation performance is measured by the intersection-over-union (IoU) score \cite{everingham2010pascal}, the pixel accuracy and the mean accuracy \cite{LongSD14}.

The first $5$ convolution blocks and the first convolution layer in the $6$th convolution block are initialized from the VGG-16 network \cite{simonyan2014very}.
All remaining layers are randomly initialized. All layers are trained using back-propagation/SGD.
As illustrated in Fig.~\ref{fig:pairwise},
we use $2$ types of pairwise potential functions.
In total, we have 1 type of unary potential function and 2 types of pairwise potential functions.
We formulate one specific FeatMap-Net and potential network (Unary-Net or Pairwise-Net) for one type of potential
function. 
We apply simple data augmentation in the training stage;
specifically, we perform random scaling (from $0.7$ to $1.2$) and flipping of the images for training.
Our system is built on MatConvNet \cite{matconvnet}.

\subsection{Results on NYUDv2}

We first evaluate our method on the dataset NYUDv2 \cite{silberman2012indoor}.
NYUDv2 dataset has 1449 RGB-D images. We use the segmentation labels provided in \cite{gupta2013perceptual} in which labels are processed into $40$ classes.
We use the standard training set which contains $795$ images and the test set which contains $654$ images.
We train our models only on RGB images without using the depth information.

Results are shown in Table \ref{tab:nyud}.
Unless otherwise specified, our models are initialized using the VGG-16 network.
VGG-16 is also used in the competing method FCN \cite{LongSD14}.
Our contextual model with CNN pairwise potentials achieves the best performance, which sets a new state-of-the-art result on the NYUDv2 dataset.
Note that we do not use any depth information in our model.

\paragraph{Component Evaluation}
We evaluate the performance contribution of different components of the FeatMap-Net for capturing patch-background context on the NYUDv2 dataset.
We present the results of adding different components of FeatMap-Net in Table \ref{tab:featmapnet}.
We start from a baseline setting of our FeatMap-Net (``FullyConvNet Baseline" in the result table), for which multi-scale and sliding pooling is removed.
This baseline setting is the conventional fully convolution network
for segmentation, which can be considered as our implementation of the FCN method in \cite{LongSD14}.
The result shows that our CNN baseline implementation (``FullyConvNet") achieves very similar performance (slightly better) than the FCN method.
Applying multi-scale network design and sliding pyramid pooling significantly improve the performance,
which clearly shows the benefits of encoding rich background context in our approach.
Applying the dense CRF method \cite{krahenbuhl2012efficient} for boundary refinement gains further improvement.
Finally, adding our contextual CNN pairwise potentials brings significant further improvement, for which we achieve the best performance in this dataset.

\begin{table}[t]
\caption{Segmentation results on NYUDv2 dataset (40 classes).
We compare to a number of recent methods.
Our method significantly outperforms the existing methods.}
\centering
\resizebox{1\linewidth}{!}
  {
  \begin{tabular}{ r | c | c c c }
method  &training data  &pixel accuracy &mean accuracy  &IoU\\ \hline \hline
Gupta et al. \cite{gupta2014learning}   &RGB-D  &60.3 &-  &28.6\\
FCN-32s \cite{LongSD14} &RGB  &60.0 &42.2 &29.2\\
FCN-HHA \cite{LongSD14} &RGB-D  &65.4 &46.1 &34.0\\ \hline
ours  &RGB  &\bf 70.0 &\bf 53.6 &\bf 40.6\\
 \end{tabular}
  }
\label{tab:nyud}
\end{table}

\begin{table}[t]
\caption{Ablation Experiments. The table shows the value added
by the different system components of our method on the NYUDv2 dataset (40 classes).
}
\centering
\resizebox{1\linewidth}{!}
  {
  \begin{tabular}{ r | c c c }
method  &pixel accuracy &mean accuracy  &IoU\\ \hline \hline
FCN-32s \cite{LongSD14} &60.0 &42.2 &29.2\\ \hline
FullyConvNet Baseline  &61.5 &43.2 &30.5\\
$+$ sliding pyramid pooling  &63.5 &45.3 &32.4\\
$+$ multi-scales  &67.0 &50.1 &37.0\\
$+$ boundary refinement &68.5 &50.9 &38.3\\
$+$ CNN contextual pairwise &70.0 &53.6 &40.6\\
 \end{tabular}
  }
\label{tab:featmapnet}
\end{table}

\begin{table*}[t]
\caption{Individual category results on the PASCAL VOC 2012 test set (IoU scores). Our method performs the best}
\centering
\resizebox{1\linewidth}{!}
  {
  \begin{tabular}{ r | c c c c c c c c c c c c c c c c c c c c | c }

method & \rot{aero}  &\rot{bike} &\rot{bird} &\rot{boat} &\rot{bottle}   &\rot{bus}  &\rot{car}  &\rot{cat}  &\rot{chair}    &\rot{cow}  &\rot{table}    &\rot{dog}  &\rot{horse}    &\rot{mbike}    &\rot{person}   &\rot{potted}   &\rot{sheep}    &\rot{sofa} &\rot{train}    &\rot{tv}  & mean \\ \hline \hline
\multicolumn{22}{c}{\bf Only using VOC training data} \\ \hline
FCN-8s \cite{LongSD14}    &76.8   &34.2   &68.9   &49.4   &60.3   &75.3   &74.7   &77.6   &21.4   &62.5   &46.8   &71.8   &63.9   &76.5   &73.9   &45.2   &72.4   &37.4   &70.9 & 55.1 & 62.2 \\
Zoom-out \cite{MostajabiYS14}  &85.6 &37.3 &\bf 83.2 &62.5 &66.0 &85.1 &80.7 &84.9 &27.2 &73.2 &57.5 &78.1 &79.2 &81.1 &77.1 &53.6 &74.0 &49.2 &71.7 &63.3 &69.6\\
DeepLab \cite{ChenPKMY14} &84.4 &54.5 &81.5 &63.6 &65.9 &85.1 &79.1 &83.4 &30.7 &74.1 &59.8 &79.0 &76.1 &83.2 &80.8 &59.7 &82.2 &50.4 &73.1 &63.7 &71.6 \\
CRF-RNN \cite{zheng2015conditional} &87.5 &39.0 &79.7 &64.2 &68.3 &87.6 &80.8 &84.4 &30.4 &78.2 &60.4 &80.5 &77.8 &83.1 &80.6 &59.5 &82.8 &47.8 &78.3 &67.1 &72.0 \\
DeconvNet \cite{noh2015learning} &89.9 &39.3 &79.7 &63.9 &68.2 &87.4 &81.2 &86.1 &28.5 &77.0 &62.0 &79.0 &80.3 &83.6 &80.2 &58.8 &\bf 83.4 &54.3 &80.7 &65.0 &72.5\\
DPN \cite{LiuDPN}  &87.7  &\bf 59.4 &78.4 &64.9 &70.3 &89.3 &83.5 &86.1 &31.7 &79.9 &\bf 62.6 &81.9 &80.0 &83.5 &82.3 &60.5 &83.2 &53.4 &77.9 &65.0 &74.1\\
ours &\bf 90.6  &37.6 & 80.0 &\bf 67.8 &\bf 74.4 &\bf 92.0 &\bf 85.2 &\bf 86.2 &\bf 39.1 &\bf 81.2 & 58.9 &\bf 83.8 &\bf 83.9 &\bf 84.3 &\bf 84.8 &\bf 62.1 & 83.2 &\bf 58.2 &\bf 80.8 &\bf 72.3 & \best 75.3 \\ \hline \hline
\multicolumn{22}{c}{\bf Using VOC+COCO training data} \\ \hline
DeepLab \cite{ChenPKMY14} &89.1 &38.3 &88.1 &63.3 &69.7 &87.1 &83.1 &85.0 &29.3 &76.5 &56.5 &79.8 &77.9 &85.8 &82.4 &57.4 &84.3 &54.9 &80.5 &64.1 &72.7 \\
CRF-RNN \cite{zheng2015conditional} &90.4 &55.3 &88.7 &68.4 &69.8 &88.3 &82.4 &85.1 &32.6 &78.5 &64.4 &79.6 &81.9 &\bf 86.4 &81.8 &58.6 &82.4 &53.5 &77.4 &70.1 &74.7\\
BoxSup \cite{Dai2015arXiv} &89.8  &38.0 &\bf 89.2 &\bf 68.9 &68.0 &89.6 &83.0 &87.7 &34.4 &83.6 &\bf 67.1 &81.5 &83.7 &85.2 &83.5 &58.6 &84.9 &55.8 &\bf 81.2 &70.7 &75.2\\
DPN \cite{LiuDPN} &89.0 & \bf 61.6 &87.7 &66.8 &74.7 &91.2 &\bf 84.3 &87.6 &36.5 & 86.3 &66.1 &84.4 &87.8 &85.6 &85.4 &63.6 &87.3 &61.3 &79.4 &66.4 &77.5 \\
ours+ &\bf 94.1	&40.7	&84.1	&67.8	&\bf 75.9	&\bf 93.4	&\bf 84.3	&\bf 88.4	&\bf 42.5	&\bf 86.4	&64.7	&\bf 85.4	&\bf 89.0	& 85.8	&\bf 86.0	&\bf 67.5	&\bf 90.2	&\bf 63.8	&80.9	&\bf 73.0	&\best 78.0 \\

\end{tabular}
  }
\label{tab:voc12_test_details}
\end{table*}

\newcounter{img_idx}

\newcounter{cntone}
\newcommand\settextone[2]{%
  \csdef{textone#1}{#2}}
\newcommand\addtextone[1]{%
  \stepcounter{cntone}%
  \csdef{textone\thecntone}{#1}}
\newcommand\gettextone[1]{%
  \csuse{textone#1}}

\newcounter{cnttwo}
\newcommand\settexttwo[2]{%
  \csdef{texttwo#1}{#2}}
\newcommand\addtexttwo[1]{%
  \stepcounter{cnttwo}%
  \csdef{texttwo\thecnttwo}{#1}}
\newcommand\gettexttwo[1]{%
  \csuse{texttwo#1}}

\addtextone{2007_001311}
\addtextone{2007_001284}
\addtextone{2007_001430}
\addtextone{2008_000149}
\addtextone{2007_007470}
\addtextone{2007_000762}
\addtextone{2008_000533}

\addtexttwo{2010_000666}
\addtexttwo{2007_000830}
\addtexttwo{2007_009346}
\addtexttwo{2009_003666}
\addtexttwo{2007_002624}
\addtexttwo{2010_005860}
\addtexttwo{2008_003333}

\newcounter{img_total_one}
\setcounter{img_total_one}{\arabic{cntone}}
\stepcounter{img_total_one}

\newcounter{img_total_two}
\setcounter{img_total_two}{\arabic{cnttwo}}
\stepcounter{img_total_two}

\begin{figure}[t]
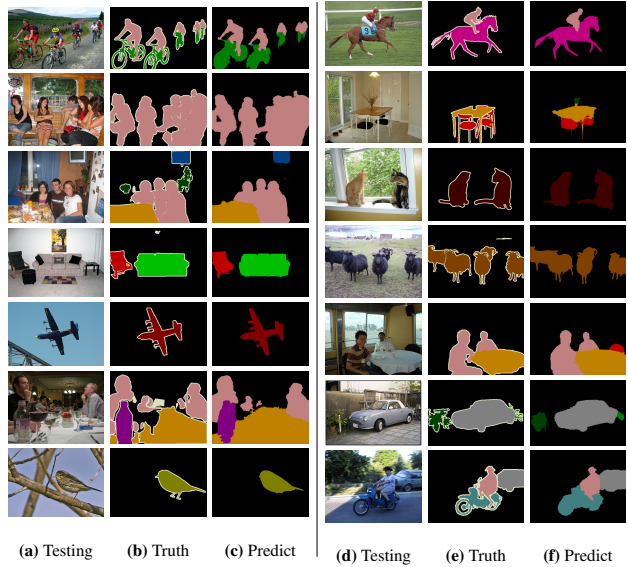

\centering
\resizebox{1\linewidth}{!} {
    \begin{subfigure}{0.7in}
    \centering
    \forloop{img_idx}{1}{\value{img_idx} < \value{img_total_one}}{
        \includegraphics[width=0.7in,height=.8in,keepaspectratio]
        {examples/voc2012_val/img/{\gettextone{\arabic{img_idx}}}.jpg}\vspace{2pt}    
    }
    \vskip -7pt
    \caption{Testing}
    \end{subfigure}\hspace{1pt}
    \begin{subfigure}{0.7in}
    \centering
    \forloop{img_idx}{1}{\value{img_idx} < \value{img_total_one}}{
        \includegraphics[width=0.7in,height=.8in,keepaspectratio]
        {examples/voc2012_val/gt/{\gettextone{\arabic{img_idx}}}.png}\vspace{2pt}    
    }
    \vskip -7pt
    \caption{Truth}
    \end{subfigure}\hspace{1pt}
    \begin{subfigure}{0.7in}
    \centering
    \forloop{img_idx}{1}{\value{img_idx} < \value{img_total_one}}{
        \includegraphics[width=0.7in,height=.8in,keepaspectratio]
        {examples/voc2012_val/predict/{\gettextone{\arabic{img_idx}}}.png}\vspace{2pt}    
    }
    \vskip -7pt
    \caption{Predict}
    \end{subfigure}\hspace{2pt}

    \hfill\vrule\hfill\hspace{2pt}

    \begin{subfigure}{0.7in}
    \centering
    \forloop{img_idx}{1}{\value{img_idx} < \value{img_total_two}}{
        \includegraphics[width=0.7in,height=.8in,keepaspectratio]
        {examples/voc2012_val/img/{\gettexttwo{\arabic{img_idx}}}.jpg}\vspace{2pt}    
    }
    \vskip -7pt
    \caption{Testing}
    \end{subfigure}\hspace{1pt}
    \begin{subfigure}{0.7in}
    \centering
    \forloop{img_idx}{1}{\value{img_idx} < \value{img_total_two}}{
        \includegraphics[width=0.7in,height=.8in,keepaspectratio]
        {examples/voc2012_val/gt/{\gettexttwo{\arabic{img_idx}}}.png}\vspace{2pt}    
     }
     \vskip -7pt
     \caption{Truth}
     \end{subfigure}\hspace{1pt}
     \begin{subfigure}{0.7in}
     \centering
     \forloop{img_idx}{1}{\value{img_idx} < \value{img_total_two}}{
        \includegraphics[width=0.7in,height=.8in,keepaspectratio]
        {examples/voc2012_val/predict/{\gettexttwo{\arabic{img_idx}}}.png}\vspace{2pt}    
     }
     \vskip -7pt
     \caption{Predict}
     \end{subfigure} 
}
\vspace{2pt}
    \caption{Some prediction examples of our method.}
    \label{fig:example_voc}
\end{figure}

\subsection{Results on PASCAL VOC 2012}

PASCAL VOC 2012 \cite{everingham2010pascal} is a well-known segmentation evaluation dataset which consists of 20 object categories and one background category.
This dataset is split into a training set, a validation set and a test set,
which respectively contain $1464$, $1449$ and $1456$ images.
Following a conventional setting in \cite{BharathECCV2014,ChenPKMY14}, the training set is augmented by extra annotated VOC images provided in \cite{HariharanABMM11}, which results in $10582$ training images.
We verify our performance on the PASCAL VOC 2012 test set.
We compare with a number of recent methods with competitive performance.
Since the ground truth labels are not available for the test set,
we report the result through the VOC evaluation server.

The results of IoU scores are shown in the last column of Table \ref{tab:voc12_test_details}.
We first train our model only using the VOC images.
We achieve $75.3$ 
IoU score which is the best result amongst methods that only use the VOC training data.

To improve the performance, following the setting in recent work \cite{ChenPKMY14,Dai2015arXiv},
we train our model with the extra images from the COCO dataset \cite{lin2014microsoft}.
With these extra training images, we achieve an IoU score of $77.2$.

For further improvement, we also exploit the the middle-layer features 
as in the recent methods \cite{ChenPKMY14,LongSD14,hariharan2014hypercolumns}.
We learn extra refinement layers on the feature maps from middle layers to refine the coarse prediction.
The feature maps from the middle layers encode lower level visual information which
helps to predict details in the object boundaries.
Specifically, we add $3$ refinement convolution layers 
on top of the feature maps from the first $5$ max-pooling layers and the input image.
The resulting feature maps and the coarse prediction score map
 are then concatenated and go through another $3$ refinement convolution layers to output the refined prediction.
The resolution of the prediction is increased from $1/16$ (coarse prediction) to $1/4$ of the input image.
With this refined prediction, we further perform boundary refinement \cite{krahenbuhl2012efficient} to generate the final prediction.
Finally,  we achieve an IoU score of $78.0$, {\em which is best reported result on this challenging dataset.}
\footnote{The result link at the VOC evaluation server: \url{http://host.robots.ox.ac.uk:8080/anonymous/XTTRFF.html}}

The results for each category are shown in Table \ref{tab:voc12_test_details}.
We outperform competing methods in most categories.
For only using the VOC training set, our method outperforms the second best method, DPN \cite{LiuDPN}, 
on $18$ categories out of $20$.
Using VOC+COCO training set, our method outperforms DPN \cite{LiuDPN}
on $15$ categories out of $20$.
Some prediction examples of our method are shown in Fig. \ref{fig:example_voc}.

\begin{table}[t]
\caption{Segmentation results on PASCAL-Context dataset (60 classes).
Our method performs the best.}
\centering
\resizebox{.8\linewidth}{!}
  {
  \begin{tabular}{ r | c c c}
  method	&pixel accuracy	&mean accuracy	&IoU\\ \hline \hline
O2P \cite{carreira2012semantic}	&-	&-	&18.1\\
CFM \cite{dai2014convolutional}	&-	&-	&34.4\\
FCN-8s \cite{LongSD14}	&65.9	&46.5	&35.1\\
BoxSup \cite{Dai2015arXiv}	&-	&-	&40.5\\ \hline
ours	&\bf 71.5	&\bf 53.9	&\bf 43.3\\
 \end{tabular}
  }
\label{tab:pascalcontext}
\end{table}

\begin{table}[t]
\caption{Segmentation results on SIFT-flow dataset (33 classes).
Our method performs the best.}
\centering
\resizebox{1\linewidth}{!}
  {
  \begin{tabular}{ r | c c c}
  method	&pixel accuracy	&mean accuracy	&IoU\\ \hline \hline
Liu et al. \cite{liu2011sift}	&76.7	&-	&-\\
Tighe et al. \cite{tighe2013finding} 	&75.6	&41.1	&-\\
Tighe et al. (MRF) \cite{tighe2013finding}	&78.6	&39.2	&-\\
Farabet et al. (balance) \cite{farabet2013learning}	&72.3	&50.8	&-\\
Farabet et al. \cite{farabet2013learning} 	&78.5	&29.6	&-\\
Pinheiro et al. \cite{pinheiro2013recurrent} &77.7 &29.8 &-\\
FCN-16s \cite{LongSD14}	&85.2	&51.7	&39.5\\ \hline
ours	&\bf 88.1	&\bf 53.4	&\bf 44.9\\
 \end{tabular}
  }
\label{tab:siftflow}
\end{table}

\subsection{Results on PASCAL-Context}
The
PASCAL-Context \cite{mottaghi2014role} dataset provides the segmentation labels of the whole scene (including the ``stuff" labels) for the PASCAL VOC images.
We use the segmentation labels which contain $60$ classes ($59$ classes plus the `` background" class ) for evaluation.
We use the provided training/test splits.
The training set contains $4998$ images and the test set has $5105$ images.

Results are shown in Table \ref{tab:pascalcontext}. Our method significantly outperforms the competing methods. To our knowledge, {\em ours is the best reported result on this dataset}.

\subsection{Results on SIFT-flow}

We further evaluate our method on the SIFT-flow dataset.
This dataset contains $2688$ images and provide the segmentation labels for $33$ classes.
We use the standard split for training and evaluation.
The training set has $2488$ images and the rest $200$ images are for testing.
Since images are in small sizes, we upscale the image by a factor of $2$ for training.
Results are shown in Table \ref{tab:siftflow}. We achieve the best performance for this dataset.

\section{Conclusions}

We have proposed a method which combines CNNs and CRFs to exploit complex
contextual information for semantic image segmentation.
We formulate CNN based pairwise potentials for modeling
semantic relations between image regions.
Our method shows best performance on several popular datasets including the PASCAL VOC 2012 dataset.
The proposed method is potentially widely applicable to other vision tasks.

\paragraph{Acknowledgments}
This research was supported by the Data to Decisions
Cooperative Research Centre and by the Australian Research Council
through the Australian Centre for Robotic Vision (CE140100016).
C. Shen's participation was supported by  an ARC Future Fellowship (FT120100969).
I. Reid's participation was supported by an ARC Laureate Fellowship (FL130100102).

C. Shen is the corresponding author (e-mail: chunhua.shen@adelaide.edu.au).

{\small
\bibliographystyle{ieee}
\bibliography{CSRef}
}

\end{document}